\documentclass[sigconf]{acmart}
\AtBeginDocument{%
  }

\setcopyright{none}
\renewcommand\footnotetextcopyrightpermission[1]{}
\usepackage{xspace}
\usepackage{multirow}
\usepackage{cuted}
\usepackage{enumitem}
\usepackage{makecell}
\usepackage[most]{tcolorbox}
\usepackage{algorithm2e}
\usepackage{listings}
\usepackage{pifont}
\usepackage{subcaption}
\theoremstyle{plain}

\theoremstyle{definition}

\theoremstyle{remark}

\newcommand{\Comment}[1]{\iffalse #1 \fi}

\newcommand{\zilong}[1]{\textcolor{orange}{\bf\small [#1 -- ZL]}}

\newcommand{\our}{Falconer\xspace}

\usepackage[textsize=tiny]{todonotes}
\newtcolorbox{userbox}[2][]{
  colback=blue!5,    
  colframe=blue!75!black, 
  title=Prompt,
  #1
}
\newtcolorbox{botbox}[2][]{%
  colback=green!10,        
  colframe=green!70!black, 
  title={#2},
  breakable,
  #1                       
} 

\begin{document}

\title{A Tale of LLMs and Induced Small Proxies: Scalable Small Language Models for Knowledge Mining}

\author{Sipeng Zhang}
\email{siz018@ucsd.edu}
\affiliation{%
  \institution{University of California, San Diego}
  \city{San Diego}
  \state{California}
  \country{USA}
}

\author{Longfei Yun}
\email{loyun@ucsd.edu}
\affiliation{%
  \institution{University of California, San Diego}
  \city{San Diego}
  \state{California}
  \country{USA}
}

\author{Zilong Wang}
\email{ziw049@ucsd.edu}
\affiliation{%
  \institution{University of California, San Diego}
  \city{San Diego}
  \state{California}
  \country{USA}
}

\author{Letian Peng}
\email{lepeng@ucsd.edu}
\affiliation{%
  \institution{University of California, San Diego}
  \city{San Diego}
  \state{California}
  \country{USA}
}

\author{Jingbo Shang}
\email{jshang@ucsd.edu}
\affiliation{%
  \institution{University of California, San Diego}
  \city{San Diego}
  \state{California}
  \country{USA}
}

\renewcommand{\shortauthors}{Sipeng et al.}
\begin{abstract}
Large language models (LLMs) exhibit strong instruction-following, planning, and reasoning capabilities, enabling new paradigms for knowledge mining---for example, Deep-Research-style extraction of labels and spans from massive corpora based on simple natural-language instructions. However, invoking LLMs at corpus scale for each classification or extraction instance is prohibitively expensive. In contrast, small models and classic pipelines are efficient but difficult to retarget and often brittle under distribution shifts. We present \our, a framework that combines the planning capability of LLMs with the efficiency of traditional data mining models (e.g., BERT). \our uses an LLM to synthesize an executable workflow and generate supervision, while offloading most inference to lightweight proxy models. These proxies expose two primitives, \texttt{get\_label} and \texttt{get\_span}, allowing a single instruction-following proxy to replace many task-specific classifiers and extractors. We further introduce new benchmarks that evaluate both planning and end-to-end knowledge mining. Across tasks, \our matches strong LLM baselines while reducing inference cost by up to 90\% and accelerating large-scale mining by over 20$\times$, making Deep-Research-style extraction practical at scale.
\end{abstract}

\ccsdesc[500]{Information systems~Structured text search}

\ccsdesc[500]{Information systems~Data mining}

\keywords{Knowledge Mining, Scalable AI}
\begin{teaserfigure}
    \centering        
    \includegraphics[width=0.98\textwidth,trim=0 180 0 180, clip]{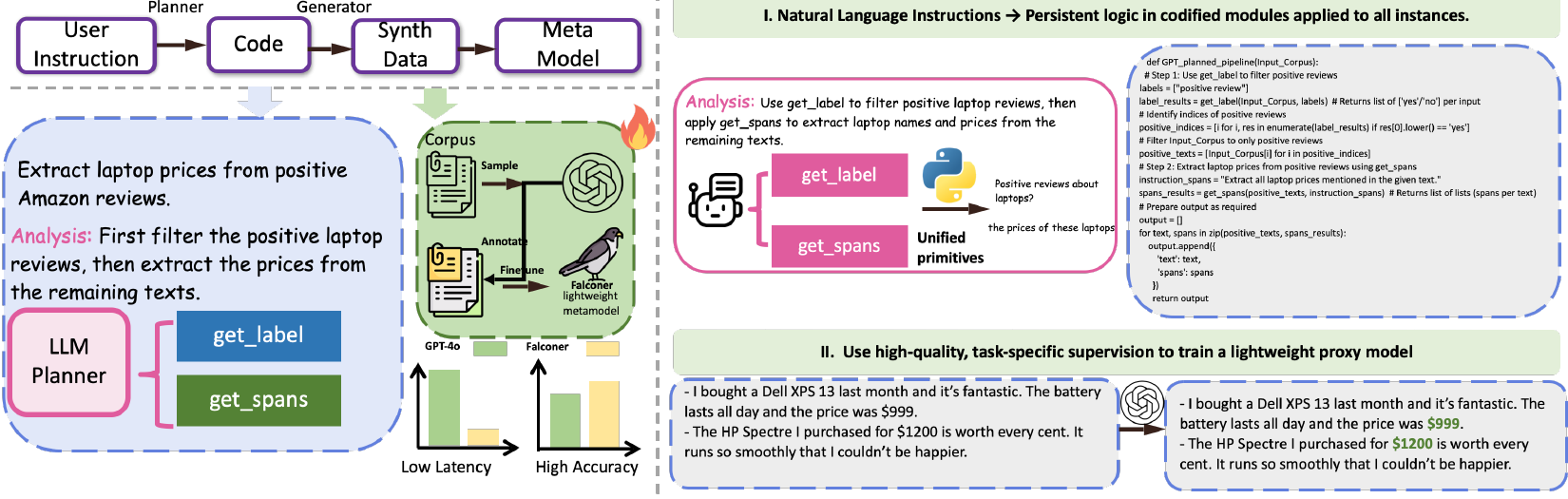}
    \vspace{-3mm}
    \caption{\our decomposes the instruction \textit{Extract all laptop prices from positive Amazon reviews} into the primitives \texttt{get\_label} and \texttt{get\_span}, generates supervision to train proxy models, and executes these primitives efficiently via small-model inference. On the right, we illustrate how \our instantiates the subtasks: it first classifies reviews as positive and then extracts the corresponding price spans. This design combines the instruction-following capability of LLMs with the efficiency of small models.}
    \label{fig:falconer_overall_workflow}
\end{teaserfigure}

\maketitle
\section{Introduction}
\begin{figure*}[htbp] 
    \centering
    \includegraphics[width=1\textwidth,trim=30 160 0 140, clip]{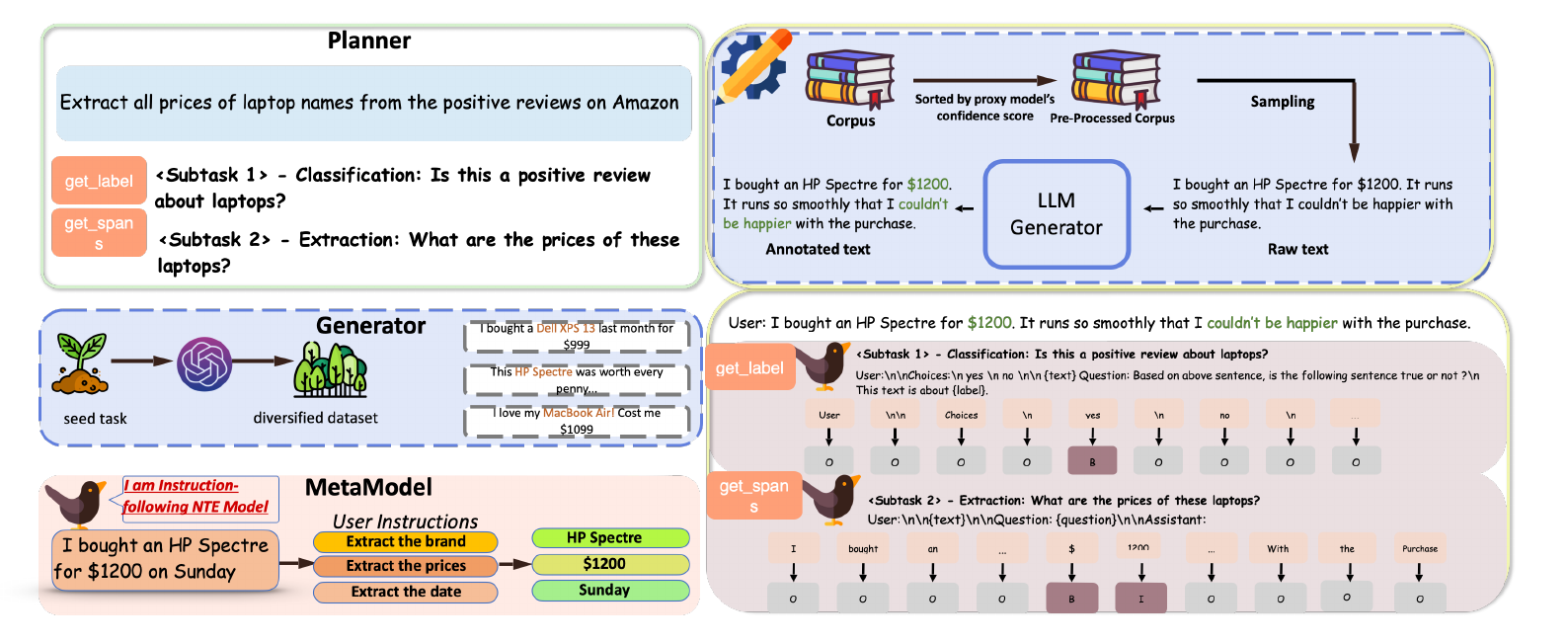} 
    \caption{Components Overview}
    \label{fig:cls_pretraining}
\end{figure*}
Knowledge mining operates over massive text corpora to extract structured information that satisfies user-defined instructions~\citep{xudocks,boylan2025glirel,wang2025speculative,ma2024star,ace2005multilingual}.
It underpins a wide range of applications in industry, science, and technology; representative examples include parsing customer reviews, analyzing biomedical literature, and summarizing large collections of technical documents~\citep{fewnerd,multinerd,TadNER,NuNER,metaie}.
Because these workloads often span millions of records, efficiency is a first-order requirement: systems must sustain high throughput and low cost while maintaining accurate and reliable outputs.

Large language models (LLMs) offer powerful instruction-following and reasoning capabilities, achieving high accuracy on knowledge mining tasks~\citep{openai-gpt5,anthropic2025claude,comanici2025gemini25,llm4clinicalie,gpt-ner,llm4ie}.
Yet using LLMs as direct executors is computationally prohibitive: each inference incurs substantial latency and cost, and scaling them to millions of documents quickly becomes infeasible.
Conversely, modern small language models (SLMs) deliver excellent inference efficiency and low deployment cost, but their limited reasoning and weak adaptability lead to poor performance on real-world knowledge mining workloads.

This mismatch motivates a hybrid design.
We introduce \our, a framework that combines the instruction understanding and decomposition strengths of LLMs with the execution efficiency of purpose-built small models.
Rather than executing knowledge mining pipelines end-to-end with LLMs, \our uses LLMs sparingly in two complementary roles.
First, as planners, they translate natural-language task descriptions into structured workflow code that specifies how SLM-based components should be invoked.
Second, as annotators, they generate supervision to train lightweight proxy models that implement the workflow efficiently at scale.
Once trained, these proxies replace LLMs as the primary executors, enabling throughput suitable for large corpora.

In \our, knowledge mining workflows are built from two primitive operations---(1) \texttt{get\_label(text, instruction)} for classification and (2) \texttt{get\_span(text, instruction)} for span extraction.
These primitives are the atomic calls that SLM-based proxies are optimized to execute.
LLM planners compose these primitives into executable pipelines.
For example, the instruction ``Extract all laptop prices from positive Amazon reviews'' becomes the workflow shown in Figure~\ref{fig:falconer_overall_workflow}.


More complex tasks, such as relation extraction, entity linking, or multi-hop queries, are expressed as similarly structured compositions.
By delegating planning and supervision to LLMs and execution to optimized small models, \our achieves both high accuracy and scalable throughput.


This abstraction enables \our to unify pipeline construction and execution within a single instruction-following proxy model, eliminating the need for separately trained, task-specific IE components and enabling training paradigms tailored to knowledge-mining workloads.
In our implementation, we use Cuckoo~\citep{peng2025cuckoo}, a lightweight instruction-following model that is the first to be pre-trained specifically for information extraction under the next-token-extraction (NTE) paradigm.
NTE allows Cuckoo to support both \texttt{get\_label} and \texttt{get\_span} within one model while following natural-language instructions without predefined label sets or schema-specific engineering.
Remarkably, despite its small size, Cuckoo delivers stronger IE performance than LLM-based extractors more than 20$\times$ larger, making it an efficient execution engine for \our and a unified, adaptive alternative to brittle, hand-crafted pipelines.


To evaluate instruction-following knowledge mining, we introduce new benchmarks that assess both planning quality and end-to-end execution. 
These benchmarks measure consistency between proxy outputs and annotations from humans and large models. 
Experiments show that while LLMs remain strong planners, their scalability is limited in practice. 
In contrast, \our closely matches state-of-the-art LLMs in accuracy while reducing inference cost by up to 90\% and accelerating large-scale processing by over 20x, making Deep Research style knowledge mining practical at scale.


In summary, our main contributions are threefold:
\begin{itemize}[nosep,leftmargin=*]
    \item We propose \our, a framework in which LLMs serve as planners and annotators: they decompose natural-language instructions into executable pipelines and generate supervision for training purpose-built lightweight proxies for knowledge mining.

    \item We introduce a specifically designed \textbf{instruction-following proxy} that unifies classification and extraction as two atomic operations (\texttt{get\_label}, \texttt{get\_span}), enabling a single small model to replace multiple task-specific components. We further show that autoregressive small models exhibit limited generalization and adaptation on knowledge-mining-style NER tasks.

    \item We construct new \textbf{instruction-following benchmarks} for knowledge mining that evaluate both planning and end-to-end execution. Experiments show that \our closely tracks state-of-the-art LLMs while reducing inference cost by up to 90\% and accelerating large-scale processing by over 20$\times$.
\end{itemize}

\vspace{-1em}
\section{Related Works}

\paragraph{Information Extraction} 
Information extraction (IE) is one of the most fundamental applications
in knowledge mining. IE systems take the user’s requirement (e.g., defined by a label text, a question, or an instruction) and extract spans of
several tokens from input texts. IE encompasses a wide range of task formulations with different level of difficulties, which varies from simple structure entity and relation extraction such as named entity recognition~\citep{sang2003introduction}, relation extraction~\citep{carreras2005introduction} , and event extraction~\citep{walker2006ace}, to more difficult tasks such as abstratc entity extraction~\citep{pontiki2016semeval, xu2020position}. 

\paragraph{LLM Agents} Recent work leverages the advanced reasoning and comprehension abilities of large language models (LLMs) to tackle diverse downstream tasks~\citep{besta2024graph,yao2023tree,shinn2023reflexion}. For complex scenarios, LLMs have been framed as autonomous agents that interact with environments~\citep{chen2023teaching,yao2023react,lu2023chameleon}, employ external tools~\citep{wu2024avatar,zong2024triad,peng2023check,durante2024agent}, and accumulate experiential knowledge~\citep{fu2024autoguide,zhao2024expel}. A representative example is ReAct~\citep{yao2023react}, which tightly integrates reasoning and action by alternating between intermediate reasoning and external operations such as information retrieval.

\paragraph{LLM Agents for Retrieval} LLM agents have been applied to Information Retrieval (IR) through pretraining, reranking, and prompting~\citep{zhuang2023open,shen2023large,wang2023query2doc}. As retrievers directly impact downstream tasks such as retrieval-augmented generation~\citep{lewis2020retrieval} and knowledge-intensive QA, domain-specific agents like EHRAgent~\citep{shi2024ehragent} have been developed to incorporate structured tool-use planning process and an interactive coding mechanism. Nevertheless, existing approaches largely depend on heuristic prompts or few-shot examples, providing limited guidance for effective retrieval strategies and tool-assisted actions.


\section{Method}
\subsection{Preliminary}
\subsubsection{Background}
Modern large language models (LLMs) exhibit strong instruction-following capabilities and generalize well across downstream tasks. However, they scale poorly for knowledge mining workloads that require processing large volumes of data. Traditional knowledge mining systems improve efficiency by chaining specialized classifiers and extractors (e.g., named entity recognition models), but they lack the instruction-following flexibility of LLMs and therefore require developers to manually construct rigid, task-specific pipelines. Recent agentic frameworks aim to balance efficiency with adaptability; nevertheless, they often rely on autoregressive small models as executors, which can limit generalization on tasks that require fine-grained, high-quality token-level supervision (e.g., knowledge mining and NER).

\subsubsection{Notations}
We briefly describe how different model families perform knowledge mining.
Our goal is to extract informative spans from an input sequence $[x_1, x_2, \ldots, x_N]$. BIO-style token classification models predict a tag sequence $[y_1, y_2, \ldots, y_N]$, where each $y_i \in \{B, I, O\}$. In contrast, autoregressive models typically generate an output sequence $[x_{N+1}, x_{N+2}, \ldots, x_{N+K}]$, where $K$ is the number of generated tokens and $x_{N+1,\,\dots,\,N+K} \in \{x_1, x_2, \ldots, x_N\}$ correspond to extracted spans.

\subsection{\our}
Our framework consists of three core components: (1) a planner that synthesizes an executable Python function to process a corpus and produce structured outputs in a user-specified format; (2) a generator that produces high-quality supervision for task adaptation; and (3) a compact proxy metamodel that nonetheless performs robustly across diverse tasks. An overview is provided in~\autoref{fig:falconer_overall_workflow}. Formally, given a user prompt $\text{Prompt}_{\text{user}}$ and a corpus $\text{Corpus}$, \our extracts text according to $\text{Prompt}_{\text{user}}$ as
\[
\text{ExtractedText} = \text{\our}(\text{Prompt}_{\text{user}}, \text{Corpus}).
\]
Our framework takes the task prompt and output specification, uses the planner to generate execution code, and then leverages the generator and metamodel to produce an adapted proxy for execution. This yields a fully automated pipeline in which users provide a corpus and obtain high-quality outputs, achieving substantial speedups and cost reductions relative to direct LLM execution while maintaining strong performance.

\begin{table}[t]
\centering
\small
\setlength{\tabcolsep}{3.2pt} 
\renewcommand{\arraystretch}{1.1}

\resizebox{\columnwidth}{!}{%
\begin{tabular}{lccccc}
\toprule
\textbf{Method} &
\textbf{Basic} &
\makecell{\textbf{Query-}\\\textbf{Based}} &
\makecell{\textbf{Multi-}\\\textbf{Entity}} &
\textbf{Misc.} &
\makecell{\textbf{Misc. w/}\\\textbf{ICL}} \\
\midrule
Falconer w/ GPT-4.1 & \textbf{0.96} & \textbf{1.00} & \textbf{1.00} & \textbf{0.21} & \textbf{0.96}\\
Falconer w/ GPT-4o  & 0.63 & 0.78 & 1.00 & 0.19 & 0.84\\
Falconer w/ Claude 3.7 Sonnet & 0.78 & 0.80 & 0.98 & 0.19 & 0.92\\
Falconer w/ GPT-4o-mini & 0.50 & 0.19 & 0.30 & 0.00 & 0.42\\
\bottomrule
\end{tabular}%
}

\caption{Planning correctness score with different LLM as planner.}
\label{tab:planning}
\end{table}

\subsection{Unified TC/IE Proxy}\label{sec:cuckooClassification}
The original Cuckoo model is specialized for information extraction (IE) tasks, including basic IE (e.g., entity and relation extraction), query-based IE, and instruction-following IE~\citep{peng2025cuckoo}. Leveraging Cuckoo's instruction-following capability, we further extend it to text classification via tailored prompt templates. Specifically, we cast text classification as a natural language inference (NLI) problem, where the model predicts whether an input sentence entails a candidate label, as illustrated in Figure~\ref{fig:cls_pretraining}.

Given a classification task that maps an input sequence $[x_1, x_2, \ldots, x_N]$ to one of $m$ classes $\{C_1, C_2, \ldots, C_m\}$, we use the instruction-following ability of BIO-tagging models and format the input as [\texttt{"Choice"}, \texttt{":"}, \texttt{"Yes"}, \texttt{"No"}, 
$x_1, x_2, \ldots, x_N$, \texttt{"SystemPrompt"}, $C$] where where $C\in \{ C_1,C_2 \ldots C_m\}$ is a candidate class and $\mathtt{SystemPrompt}$ is a fixed template (e.g., \texttt{"Question: Based on the above sentence, is the following sentence true or not? This text is about"}).
To this end, we construct an instruction-based prompt template for classification and fine-tune Super Rainbow Cuckoo on the datasets introduced in~\citet{laurer2023building}, yielding the metamodel used in our experiments.


\begin{figure*}[htbp]
    \centering
    \begin{minipage}[h]{1\columnwidth}
        \centering
        \resizebox{\textwidth}{!}{%
            \begin{tabular}{llcccccc}
                \toprule
                \textbf{Metamodel} & \textbf{Dataset} & \textbf{64 Samples} & \textbf{512 Samples} & \textbf{GPT-4o \Comment{zero shot}} \\
                \midrule
                Cuckoo & Biology & 0.42 & \textbf{0.45} & 0.27 \Comment{0.27}\\
                RoBERTa-Large & Biology & 0.00 & 0.41 & 0.27 \\
                Qwen3-0.6B & Biology & 0.08(0.10) & 0.23(0.26) & 0.27 \\
                Gemma3-1B-It & Biology & 0.08(0.12) & 0.12(0.29) & 0.27 \\
                \midrule
                Cuckoo & Twitter & 0.19 & \textbf{0.43} & 0.35 \Comment{0.10}\\
                RoBERTa-Large & Twitter & 0.00 & 0.38 & 0.35 \\
                Qwen3-0.6B & Twitter & 0.06(0.12) & 0.28(0.33) & 0.35 \\
                Gemma3-1B-It & Twitter & 0.07(0.10) & 0.20(0.32) & 0.35 \\
                \midrule
                Cuckoo & Fabrication & 0.20 & 0.32 & \textbf{0.38} \Comment{0.12}\\
                RoBERTa-Large & Fabrication & 0.00 & 0.17 & 0.38 \\
                Qwen3-0.6B & Fabrication & 0.04(0.04) & 0.12(0.15) & 0.38 \\
                Gemma3-1B-It & Fabrication & 0.00(0.00) & 0.09(0.13) & 0.38 \\
                \midrule
                Cuckoo & Wiki & 0.03 & \textbf{0.68} & 0.53 \Comment{0.00}\\
                RoBERTa-Large & Wiki & 0.00 & 0.60 & 0.53 \\
                Qwen3-0.6B & Wiki & 0.00(0.00) & 0.57(0.55) & 0.53 \\
                Gemma3-1B-It & Wiki & 0.00(0.02) & 0.43(0.45) & 0.53 \\
                \midrule
                Cuckoo & Vehicle & 0.42 & 0.75 & \textbf{0.76} \Comment{0.24}\\
                RoBERTa-Large & Vehicle & 0.00 & 0.66 & 0.76 \\
                Qwen3-0.6B & Vehicle & 0.00(0.00) & 0.60(0.63) & 0.76 \\
                Gemma3-1B-It & Vehicle & 0.00(0.00) & 0.42(0.55) & 0.76 \\
                \bottomrule
            \end{tabular}
        }
        \captionof{table}{Results on NER Datasets with Ground Truth labels. continually pretrained models' results are given in parentheses}
        \label{tab:main_1}
    \end{minipage}
    \hfill
    \begin{minipage}[h]{1\columnwidth}
        \centering
        \includegraphics[width=\columnwidth]
        {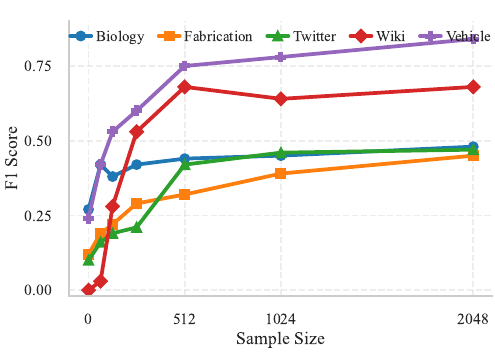}
        \captionof{figure}{Model Performance under different sample size }
        \label{fig:SampleSizeCurve}
    \end{minipage}
    \vspace{-3mm}
\end{figure*}
\subsection{Planning}

The planner is the core of \our, translating natural language requirements into executable pipelines by codifying instructions into atomic operations and explicit control flows. For a knowledge mining objective, it decomposes the input into subtasks (e.g., classification, span extraction), each bound to a tool interface such as \texttt{get\_label} or \texttt{get\_span}. These are then assembled into a deterministic control flow, ensuring explicit execution without reliance on implicit reasoning. Specifically, we prompt modern, high-capacity large language models to explicitly decompose each task into two operations \texttt{get\_label} and \texttt{get\_span}, which are internally instantiated by the proxy model, as illustrated in Figure ~\ref{fig:cls_pretraining}. Sample code is shown in Figure~\ref{fig:falconer_overall_workflow} and Appendix~\ref{code:SampleCode}.

Crucially, the planner does not merely synthesize runnable code but codifies the logical dependencies among subtasks. For example, in a multi-entity extraction scenario, \textit{Retrieve all talks about both health and brain, then extract their lecturers}, the planner constructs a sequential program where the input texts are first filtered using two classification heads for “health” and “brain,” then conditionally passed into a span extractor to identify lecturer names. This approach integrates boolean logic, ordered execution, and parameterized prompt templates into a unified representation, ensuring that downstream behavior is both interpretable and reusable across tasks.

By explicitly codifying instructions into executable task pipelines, \our provides two key benefits. First, the structured representation enables generalization across diverse task formulations, including multi-label classification and multi-entity extraction. Second, codification improves transparency: each decision can be traced back to a deterministic plan, bridging the gap between user intent and system actions.

\autoref{tab:planning} compares the planning abilities of different models. We observe that GPT-4.1 achieves high accuracy across diverse tasks, making it a strong candidate for our planner. However, performance drops on complex tasks, which we define as multi-step tasks that require intermediate execution results rather than a single fixed string (e.g., first extracting a lecturer’s name, then identifying that lecturer’s profession). To further probe model limits, we include a set of miscellaneous tasks specifically designed to stress-test state-of-the-art LLMs under such challenging scenarios. While models struggle in these cases, their accuracy improves substantially with in-context learning (ICL), underscoring both the difficulty of complex tasks and the effectiveness of our framework in decomposing knowledge mining objectives into well-structured subtasks.

\begin{table*}[htbp]
  \centering
\small
\resizebox{\textwidth}{!}{%
    \begin{tabular}{cllccccccccccc}
    \toprule
    \multirow{2}[4]{*}{} & \multirow{2}[4]{*}{\textbf{Model}} & \multirow{2}[4]{*}{\textbf{Dataset}} & \multicolumn{4}{c}{\textbf{Basic Task}} & \multicolumn{3}{c}{\textbf{Query-based Task}} & \multicolumn{4}{c}{\textbf{Multi-entity Task}} \\
\cmidrule(lr){4-7} \cmidrule(lr){8-10} \cmidrule(lr){11-14}
&       &       & \textbf{Task 1} & \textbf{Task 2} & \textbf{Task 3} & \textbf{Average} & \textbf{Task 1} & \textbf{Task 2} & \textbf{Average} & \textbf{Task 1} & \textbf{Task 2} & \textbf{Task 3} & \textbf{Average} \\
    \midrule
    \multirow{3}[2]{*}{\rotatebox{90}{\textbf{0-shot}}} & Cuckoo & TED   & 0.489 & 0.654 & 0.514 & 0.552 & 0.383 & 0.371 & 0.377 & 0.395 & 0.607 & 0.497 & 0.500 \\
          & Cuckoo & Steam Game & 0.501 & 0.683 & 0.535 & 0.573 & 0.374 & 0.350 & 0.362 & 0.451 & 0.524 & 0.468 & 0.481 \\
          & Cuckoo & Text Message & 0.584 & 0.694 & 0.585 & 0.621 & 0.418 & 0.392 & 0.405 & 0.564 & 0.583 & 0.530 & 0.559 \\
    \midrule
    \multirow{6}[2]{*}{\rotatebox{90}{\textbf{Few-shot}}} & Cuckoo & TED   & 0.658 & 0.758 & 0.683 & \textbf{0.699} & 0.532 & 0.557 & \textbf{0.545} & 0.644 & 0.692 & 0.661 & \textbf{0.666} \\
          & Roberta-Large & TED   & 0.552 & 0.587 & 0.553 & 0.564 & 0.446 & 0.511 & 0.479 & 0.517 & 0.566 & 0.531 & 0.538 \\
          & Qwen3-0.6B & TED   & 0.503 & 0.599 & 0.492 & 0.531 & 0.372 & 0.489 & 0.431 & 0.497 & 0.424 & 0.501 & 0.474 \\
          & Cuckoo & Steam Game & 0.672 & 0.783 & 0.675 & \textbf{0.710} & 0.569 & 0.587 & \textbf{0.578} & 0.673 & 0.719 & 0.684 & \textbf{0.692} \\
          & Roberta-Large & Steam Game & 0.509 & 0.525 & 0.517 & 0.517 & 0.434 & 0.452 & 0.443 & 0.588 & 0.383 & 0.564 & 0.512 \\
          & Qwen3-0.6B & Steam Game   & 0.486 & 0.521 & 0.466 & 0.491 & 0.379 & 0.423 & 0.401 & 0.524 & 0.301 & 0.492 & 0.439 \\
          & Cuckoo & Text Message & 0.703 & 0.806 & 0.731 & \textbf{0.747} & 0.590 & 0.614 & \textbf{0.602} & 0.709 & 0.726 & 0.734 & \textbf{0.723} \\
          & Roberta-Large & Text Message & 0.553 & 0.621 & 0.590 & 0.588 & 0.496 & 0.518 & 0.507 & 0.548 & 0.570 & 0.574 & 0.564 \\
          & Qwen3-0.6B & Text Message   & 0.536 & 0.587 & 0.594 & 0.571 & 0.479 & 0.476 & 0.463 & 0.532 & 0.511 & 0.525 & 0.523 \\
    \bottomrule
    \end{tabular}%
    }
    \caption{Results from various proposed tasks on 3 datasets with subtasks}
    \label{tab:t1}
\end{table*}%
\subsection{Generator}
A major challenge in adapting a lightweight metamodel to diverse knowledge mining tasks is acquiring high-quality, task-specific supervision without incurring prohibitive costs. In \our, we address this challenge with a generator that bridges raw corpus data and the specialized capabilities required by the metamodel. Unlike fully synthetic data produced by large language models, which can diverge from the target distribution, our generator leverages naturally occurring corpus examples to produce realistic, task-aligned supervision.

The generator operates in three stages. First, around five percent of the entire corpus is sampled to capture the authentic distribution of the domain, which is detailed in Appendix~\ref{sec:sampling}. Second, a powerful large language model (e.g., GPT-4.1) annotates these samples according to the planner’s codified task descriptions, covering subtasks such as entity extraction, classification, and relation detection. Importantly, the generator enriches naturally occurring data with high-quality labels rather than fabricating artificial inputs, ensuring statistical fidelity to the corpus. Finally, the annotated samples are used to fine-tune the metamodel, enabling it to acquire task-specific knowledge while maintaining its efficiency advantages over large models. A summary of performance gains is provided in Table~\ref{tab:t2}.

In subsequent experiments, this approach demonstrates high efficiency, achieving performance comparable to or even surpassing state-of-the-art large language models while using only 5\% of the original corpus. Crucially, the generator’s success hinges on access to high-quality supervision, which can be readily extended to alternative sources such as carefully curated human annotations.



\subsection{Metamodel: Lightweight Yet Capable Proxy}
In \our, the metamodel serves as the central execution engine, acting as a lightweight proxy for large language models in downstream knowledge mining tasks.
Instead of relying on general-purpose LLMs for every request, we adopt Cuckoo~\citep{peng2025cuckoo}, which has a parameter scale comparable to RoBERTa~\citep{liu2019roberta}, to strike a balance between efficiency and capability. This design enables \our to achieve near-LLM performance with substantially reduced inference cost.

Moreover, Falconer’s modular architecture leverages Cuckoo not as a monolithic generalist, but as a specialized executor within a planner-driven pipeline. The planner codifies user intents into explicit, interpretable subtasks; the metamodel then executes these subtasks with high efficiency. This separation enables Falconer to exploit the SLM-first paradigm advocated by recent research~\citep{belcak2025small}.

Empirically, this design achieves substantial gains in both efficiency and scalability. Cuckoo requires up to 20× fewer FLOPs and 1000× less memory than GPT-class models, while maintaining competitive accuracy on instruction-following and span-extraction benchmarks relevant to knowledge mining. This efficiency enables \our to operate cost-effectively across massive corpora, supporting real-time inference even in resource-constrained environments.

\section{Experiments}

We evaluate \our on a broad spectrum of knowledge mining tasks to demonstrate that a lightweight metamodel, when coupled with our planner–generator–executor framework, can achieve performance comparable to state-of-the-art LLMs while being significantly more efficient. Our experiments are designed to answer two central questions:
\begin{itemize}[itemsep=0.5pt,leftmargin=*]
    \item whether these metamodels maintain high alignment with human annotations on labeled datasets and
    \item whether \our can generate metamodels that faithfully approximate the behavior of large models(its annotator) on unlabeled corpora
\end{itemize}

All experiments reported in this section were conducted using a metamodel fine-tuned on 5\% of the original corpus annotated by an LLM, unless otherwise specified. Model performance is evaluated using the word-level F1 score.



\subsection{Labeled Dataset}

This set of experiments is primarily intended to assess the consistency between the metamodel and human annotations, as well as to benchmark the performance of the metamodel against that of contemporary large language models. Furthermore, we utilized several widely adopted Named Entity Recognition (NER) datasets, including \textbf{FabNER}, \textbf{Broad Twitter}, \textbf{BC2GM}, \textbf{AnatEM}, \textbf{WikiNER}, and \textbf{FindVehicle}. These datasets were combined to construct a new benchmark, which was subsequently employed to assess the metamodel’s performance across diverse groups of tasks. For particularly large datasets, such as WikiNER, we randomly sampled a subset to the mixed dataset. Meanwhile, to more explicitly illustrate the adaptability of the metamodel to downstream tasks, we present experimental results obtained by fine-tuning the metamodel with varying amounts of training data, ranging from 64 to 2048 samples. It is worth noting that even the largest setting of 2048 samples corresponds to only 5\% of the original corpus. The main results are shown in Table \ref{tab:main_1} and detailed results are plotted in Figure \ref{fig:SampleSizeCurve}.

From the experimental results, we observe a consistent improvement in test performance as the sample size increases. Notably, the model fine-tuned with 2048 samples \textbf{outperforms GPT-4o across all task categories}, providing strong evidence of its substantial adaptability to knowledge mining tasks. Meanwhile, the rate of performance gains is closely tied to the quality of annotations generated by the large model. When the annotations are of high quality, the metamodel tends to achieve performance saturation more rapidly, as illustrated by the experiments on WikiNER. Conversely, in tasks where the large model produces suboptimal annotations, the performance of the metamodel improves more gradually, thereby reflecting the core principle of co-evolution between the metamodel and large models ~\citep{peng2025cuckoo}. Finally, we also conduct experiments in controlled environments \textbf{comparing Falconer with modern agentic knowlegde mining framework} which mainly utilizes small language models as executor models (e.g., Qwen3-0.6B and Gemma-1B-IT). We observe that these modern autoregressive small language models show little or no improvement over target tasks. The results demonstrate that task-specialized model designs are substantially more effective at handling knowledge tasks. More details and analysis are given in Section~\ref{sec:analysis_1}



\subsection{Unlabeled Dataset Evaluation}

To evaluate the effectiveness of Falconer in generating reliable proxy metamodels, we measure the consistency scores between the metamodel and GPT-4o across three large-scale unlabeled corpora, \textbf{TED Talk Summary}\Comment{\footnote{\href{https://huggingface.co/datasets/gigant/ted_descriptions}{Huggingface: chirunder/gigant/ted\_descriptions}}}, \textbf{Steam Game Description}\Comment{\footnote{\href{https://huggingface.co/datasets/FronkonGames/steam-games-dataset}{Huggingface: FronkonGames/steam\-games\-dataset}}}, \textbf{and Text Message}\Comment{\footnote{\href{https://huggingface.co/datasets/chirunder/text_messages}{Huggingface: chirunder/text\_messages}}}. We design a diverse set of knowledge mining tasks spanning three categories: \textbf{basic tasks} involving entity recognition and simple classification, \textbf{query-based tasks} requiring sentence-level semantic understanding, and \textbf{multi-label/multi-entity task}s that demand compositional reasoning. Please refer to the complete list of tasks provided in Appendix~\ref{sec:proposed_task}

\paragraph{Basic Task} This category benchmarks the fundamental capacity of models to discern labels, entities, and relations. We construct a suite of tasks that closely approximate real-world knowledge mining settings, exemplified by sample 1 and 2 in Appendix \ref{sec:sample_task}. The task set spans elementary classification, entity and relation extraction, as well as composite formulations integrating both. For pairwise relation extraction tasks, we further stipulate that one entity participating in the relation is pre-specified, thereby isolating the model’s ability to infer the remaining relational structure. As shown in Table~\ref{tab:t1}, the tasks categorized as Basic Task demonstrate that, after fine-tuning, the metamodel consistently achieves high agreement with the large model.
\paragraph{Query-Based Task} This category of tasks focuses on assessing the model’s ability to capture more complex sentence-level semantics, as exemplified by sample 3 and 4 in Appendix \ref{sec:sample_task}. Illustrated in Table \ref{tab:t1}, the corresponding tasks are represented by Query-based Task. With appropriate fine-tuning, the metamodel demonstrates competitive performance on complex tasks. It is worth noting that the untuned metamodel exhibits the weakest performance in this category; however, fine-tuning yields substantial improvements. For instance, given the task prompt “retrieve all texts that are primarily about medicine, and extract what the lecturer will talk about”, the initial metamodel achieves an F1 score of only 0.23 when compared against GPT-4o as the reference. After fine-tuning with only a small fraction of the annotated corpus, its F1 score increases to 0.56. These results highlight the model’s strong capacity to adapt effectively to downstream tasks.
\paragraph{Multi-entity Task} 
This category of tasks extends metamodel evaluation to multi-label and multi-entity scenarios (sample 5 in Appendix \ref{sec:sample_task}). Prior work highlights the limitations of large language models in multi-label classification~\citep{ma2025large,xu2024llms}. In contrast, our framework employs the planner to decompose such tasks into sequential subtasks, whose outputs are aggregated to form the final result. For instance, the query “retrieve all speeches concerning both health and the brain” is decomposed into two classification subtasks—health-related and brain-related—whose results are combined via Boolean logic. This structured decomposition enables logically consistent and accurate performance in multi-label classification and multi-entity extraction.

The experimental results for Multi-entity Task, as reported in Table \ref{tab:t1}, indicate that the adapted metamodel demonstrates strong proficiency in handling multi-entity tasks, achieving performance that is competitive with, and in some cases surpasses, results obtained through multi-turn prompting augmented with human annotations.


\section{Analysis}

\subsection{Ablation on MetaModel}\label{sec:analysis_1}
In our experiments, we observe that pipelines employing modern state-of-the-art small language models as executor models exhibit significantly worse generalization performance on downstream tasks compared to pipelines based on small models trained from RoBERTa architecture and the Next Token Extraction(NTE) paradigm\cite{peng2025cuckoo}. More generally, we hypothesize that autoregressive models are not an optimal choice for general-purpose knowledge mining tasks, although it is always the case in other modern agentic framework\cite{hong2023metagpt,guan2024amor}. To further investigate the behavior of autoregressive models in knowledge mining, we continually training Qwen3-0.6B and Gemma-1B-IT on the same NER pretraining dataset as Cuckoo, obtaining a new set of models. These datasets are labeled with target spans typical of NER tasks and encompass diverse domains, including biology, dialogue, healthcare etc which provide informative signals for knowledge mining tasks. As shown in Table \ref{tab:main_1}, the continually pretrained models obtain little or no improvement over evaluation tasks. This suggests that autoregressive models are not effective at extracting knowledge from NER-style supervision in the training data, highlighting the need for task-specialized designs.

\subsection{Continual Integration Analysis}
\begin{figure}[t]
    \centering
    \includegraphics[width=\columnwidth]{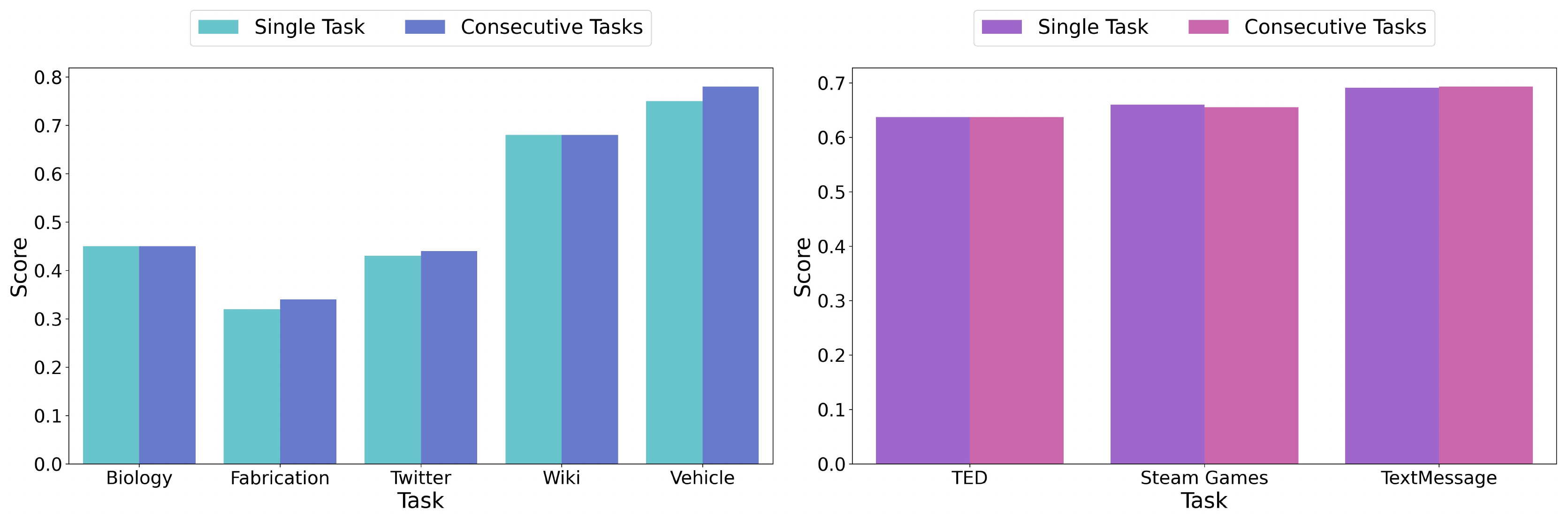}
    \caption{
    Performance on labeled datasets for single-task training with a new metamodel
    and consecutive-task training with the metamodel from the previous task.
    Performance on unlabeled datasets under the same settings.
    }
    \label{fig:continue_integration}
\end{figure}
\begin{figure}[t]
    \centering
    \includegraphics[width=\columnwidth]{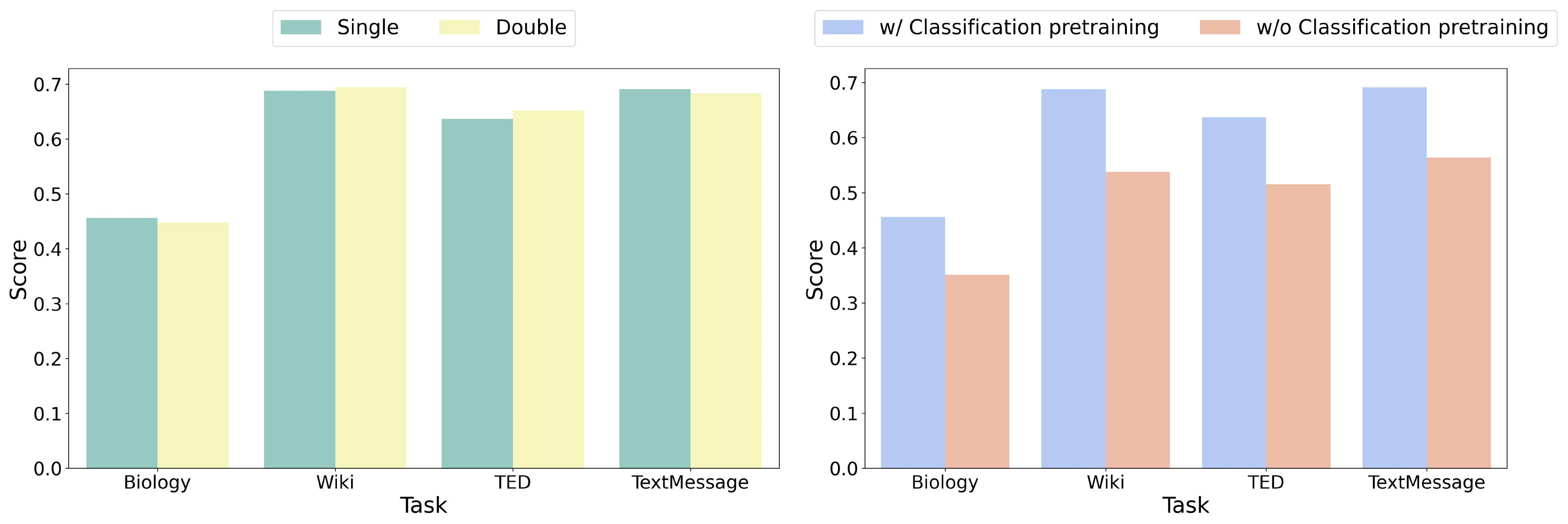}
    \caption{
    Performance with varying numbers of metamodels across different task types,
    and comparison of different pretraining strategies.
    }
    \label{fig:efficiency_analysis}
\end{figure}
\begin{figure}[htbp]
    \vspace{-0.7em}
        \begin{tabular}{lcc}
            \toprule
            \textbf{Models} & \textbf{F1 score}  \\ 
            \midrule
            Roberta-Large (degraded data) & 0.24  \\
            Roberta-Large (original data) & 0.40 \\
            Cuckoo (degraded data) & 0.41 \\
            Cuckoo (original data) & 0.42  \\
            Annotator (GPT-4o) & 0.27 \\
            \bottomrule
        \end{tabular}
        \caption{Results of different models on Biology Task}
        \label{tab:results}
    \vspace{-0.7em}
\end{figure}
In Table \ref{tab:t2}, fine-tuning was restarted from a fresh base model for each task. In practice, however, continual learning is equally important, as models are expected to sustain performance across sequential tasks while retaining competence from earlier ones. To evaluate this ability, we reformulated the setup into a sequence of five tasks, where each task used the model fine-tuned on its predecessor as the base. We report the results in Figure \ref{fig:continue_integration}, averaging over subtasks when applicable. The figure demonstrates the metamodel’s performance under sequential fine-tuning and evaluation on consecutive tasks.


We observe that models undergoing multiple rounds of fine-tuning on sequential tasks maintain capabilities comparable to those fine-tuned directly from the base model. Overall, our evaluation highlights the metamodel’s continual integration ability, demonstrating its effectiveness in sustaining high performance across a broad spectrum of real-world tasks. Moreover, the results validate that the proposed framework substantially alleviates the deployment overhead associated with adapting models to diverse tasks.


In this section, we further highlight the efficiency and performance advantages of our framework. While prior experiments benchmarked RoBERTa-large against multiple baselines, its lack of inherent instruction-following ability required training two task-specific variants for classification and extraction. By contrast, our framework enables the incubation of a single metamodel that leverages instruction-following to generalize across heterogeneous tasks. To validate this, we additionally trained two separate metamodels—one for classification and one for extraction—on the same benchmarks. As shown in left panel of Figure \ref{fig:efficiency_analysis}, their performance is nearly indistinguishable from that of a unified model, underscoring that a single metamodel can achieve state-of-the-art performance across task types while significantly reducing deployment overhead.



Meanwhile, to further substantiate the metamodel’s capacity for continual generalization across novel tasks, we additionally evaluate its performance without pretraining on the classification dataset (detailed in Section \ref{sec:cuckooClassification}). This comparison highlights the model’s adaptability, demonstrating its ability to rapidly generalize to unseen tasks through a combination of pretraining and fine-tuning. As shown in right panel of Figure \ref{fig:efficiency_analysis}, we fine-tune the model on datasets of equal size and train for the same number of epochs to ensure a controlled setting. The results indicate that the pretrained metamodel achieves significantly faster convergence when adapted to new tasks, underscoring its strong generalization and adaptability in continual learning scenarios.


\subsection{Case Study: Arising Abilities of model}\label{sec:arising}
\begin{tcolorbox}[
  title=Task: Extract all gene names in the give text,
  colback=white,              
  colframe=blue!40!black,     
  colbacktitle=blue!20,      
  coltitle=black,            
  fonttitle=\bfseries,
  label={box:arising}
]
In the course of Hepatitis A HBs - and HBe - antigen as well as HBc ( IgM and IgG ) - , HBs - and HBe - antibodies can be detected .
\tcblower
\textbf{Answers:}

\textbf{GPT-4o}:['None']~~~~~~~~~~~~~~\textbf{Untuned model}:['None']

\textbf{Tuned model}:['HBs', 'HBe', 'HBc']
\end{tcolorbox}

Table \ref{tab:t2} reveals the model’s strong performance across tasks, with notable patterns emerging. On \textbf{Biology} tasks, GPT-4o achieves an average F1 of 0.27—barely matching the metamodel’s zero-shot performance—highlighting the low quality of GPT-4o annotations. Intriguingly, fine-tuning the metamodel on these noisy labels still yields substantial gains. Manual analysis attributes \textbf{74\%} of this improvement to the phenomenon illustrated in \ref{box:arising}, which we term \textbf{arising abilities}.

As shown in \ref{box:arising}, we define arising abilities as \textbf{the model’s capacity to correct its outputs even when provided with inaccurate annotation guidance from contemporary LLMs}. Similar phenomena have been observed in prior studies ~\citep{shao2025spurious, ye2025limo}, which report that models can self-correct under random or deliberately misleading guidance. These works attribute this capability to the elicitation of the model’s extensive pretrained knowledge, aligning with our analytical interpretation. To further validate this hypothesis, we conducted a series of controlled experiments, detailed below.

The metamodel’s pretraining on IE tasks \cite{peng2025cuckoo}, which encode entities with positional information, appears to endow it with a strong sensitivity to token structure. We hypothesize that this enables the model to spontaneously extract entities at corresponding positions when faced with new entities sharing similar positional patterns. To test this, we degraded GPT-annotated data by randomizing span start positions while preserving span endings, retaining primarily positional cues. For instance, a consecutive span: \textit{cytochrome P-450 monooxygenase} could be annotated as \textbf{cytochrome} P-450 monooxygenase, cytochrome \textbf{P-450} monooxygenase or cytochrome P-450 \textbf{monooxygenase}. Fine-tuning on this degraded data yielded performance nearly identical to training on the original annotations, whereas RoBERTa-large suffered a substantial drop (Table \ref{tab:results}). These results suggest that the model’s arising ability is driven almost entirely by positional supervision, revealing a striking capability arising from its pretraining knowledge.

\section{Conclusion}
This paper proposes a framework for the automated execution of knowledge mining tasks, which decomposes each task into several subtasks and employs a unified model to perform them. Consequently, users only need to provide a task prompt and specify the output format to effortlessly execute a wide range of knowledge mining tasks, while benefiting from performance surpassing that of even the most power modern large language models, as well as 90\% inference costs decrease and 20x inference speed increase.


\bibliography{reference}
\bibliographystyle{ACM-Reference-Format}

\appendix
\section{Cuckoo For Text classification}\label{sec:classificationPrompt}


\begin{userbox}[label={prompt_template}]

User:\\
Choices: \\ yes \\ no \\
\textbf{Input Text} Question: Based on above sentence, is the following sentence true or not ? \\
This text is about \textbf{label} \\
Assistant:\\
Answer:
\end{userbox}
We adopt the aforementioned template and leverage the token-level supervision provided by Cuckoo to reformulate the classification task into a more general natural language inference (NLI) problem. An illustrative example is provided in Figure~\ref{fig:cls_pretraining}.


\section{Generating Fine-tuning Samples}\label{sec:sampling}
We leverage the metamodel’s inherent pretraining knowledge and adopt a heuristic approach to obtain a relatively high-quality fine-tuning dataset. For classification tasks, the generation of fine-tuning samples is illustrated in Algorithm~\ref{algo:heuristic}, whereas for extraction tasks, we directly employ random sampling.

\begin{algorithm}[htbp]
\caption{Classification Training Set Generation}
\label{algo:heuristic}
\KwIn{Corpus $\mathcal{C}$, label $l$, sample size $N$}
\KwOut{Training set $\mathcal{T}$}

Initialize empty set $\mathcal{T}$\;
\ForEach{sample $x \in \mathcal{C}$}{
    Compute score $s(x, l)$ using metamodel\;
}
Sort all samples in $\mathcal{C}$ by score $s(x, l)$ in descending order\;
Select top $N$ samples $\{x^+_1, \dots, x^+_N\}$ as positive set $\mathcal{P}$\;
Select bottom $N$ samples $\{x^-_1, \dots, x^-_N\}$ as negative set $\mathcal{N}$\;
Construct training set $\mathcal{T} = \mathcal{P} \cup \mathcal{N}$\;
\Return $\mathcal{T}$\;

\end{algorithm}

\section{Continual Pretraining on autoregressive models}\label{sec:more_results_on_autoregressive_models}

\begin{figure}[htbp]
    \centering
    \begin{minipage}[h]{0.48\textwidth}
        \centering
        \resizebox{\textwidth}{!}{%
            \begin{tabular}{llcccccc}
                \toprule
                \textbf{Metamodel} & \textbf{Dataset} & \textbf{64 Samples} & \textbf{512 Samples} & \textbf{GPT-4o \Comment{zero shot}} \\
                \midrule
                Qwen3-0.6B & Biology & 0.08 & 0.258 & 0.27 \\
                Gemma3-1B-It & Biology & 0.08 & 0.12 & 0.27 \\
                \midrule
                Qwen3-0.6B & Twitter & 0.06 & 0.28 & 0.35 \\
                Gemma3-1B-It & Twitter & 0.07 & 0.20 & 0.35 \\
                \midrule
                Qwen3-0.6B & Fabrication & 0.04 & 0.55 & 0.38 \\
                Gemma3-1B-It & Fabrication & 0.00 & 0.09 & 0.38 \\
                \midrule
                Qwen3-0.6B & Wiki & 0.00 & 0.55 & 0.53 \\
                Gemma3-1B-It & Wiki & 0.00 & 0.43 & 0.53 \\
                \midrule
                Qwen3-0.6B & Vehicle & 0.00 & 0.60 & 0.76 \\
                Gemma3-1B-It & Vehicle & 0.00 & 0.42 & 0.76 \\
                \bottomrule
            \end{tabular}
        }
        \captionof{table}{Results on NER Datasets with Ground Truth labels \Comment{\zilong{what does it mean by 64/512 in the table? Add a brief notation in the table to make the table self-contained.}}}
        \label{tab:t2}
    \end{minipage}
    \hfill
\end{figure}

\section{Sample Task}\label{sec:sample_task}
\begin{botbox}{Sample Task}

1. retrieve all speaks which is mainly about finance and extract its lecturer

2. extract all locations mentioned in the text

3. find all talks that address breaking gender stereotypes in modern society, and include all countries mentioned

4. retrieve all speaks which is mainly about how mental health influences our daily lives and extract all the institution name mentioned

5. retrieve all speaks which is mainly about both health and brain in the speak, then extract their lecturer

\end{botbox}

\section{Sample Planning Code}
{\scriptsize
\begin{lstlisting}[
  language=Python,
  caption={},
  label={code:SampleCode},
  breaklines=true,
  breakatwhitespace=true,
  columns=fullflexible
]
def GPT_pipeline(Input_Corpus):
    labels = ['finance']
    label_results = get_label(Input_Corpus, labels)  

    finance_indices = [i for i, result in enumerate(label_results) if result[0].lower() == 'yes']
    filtered_texts = [Input_Corpus[i] for i in finance_indices]
    if not filtered_texts:
        return []
    instruction_spans = "Extract the lecturer of the speak in the given text."
    spans_results = get_spans(filtered_texts, instruction_spans)
    output = []
    for idx, orig_idx in enumerate(finance_indices):
        output.append({
            'text': Input_Corpus[orig_idx],
            'spans': spans_results[idx]
        })
    return output
\end{lstlisting}
}



\section{Human Proposed Task on Unlabeled Datasets}\label{sec:proposed_task}
\begin{botbox}[label={all_tasks}]{Tasks on TED description Dataset}
1.retrieve all speaks which is mainly about finance and extract its lecturer \\
2.output all speaks which is mainly about mental health and extract its speakers \\
3.return all speaks which is mainly about environment and extract all the locations mentioned in the text \\
4.retrieve talks whose main theme is artificial intelligence and list all professions mentioned \\
5.get all talks that center on medicine and identify all disease mentioned \\
6.collect all speaks which is mainly about finance \\ 
7.give out all speaks which is mainly about health \\
8.retrieve all speaks which is mainly about education \\
9.gather all speaks which is mainly about technology \\
10.output all speaks which is mainly about politics \\
11.Extract all locations mentioned \\
12.Extract all time mentioned \\
13.Extract all countries mentioned \\
14.Extract all website mentioned \\
15.Extract all person mentioned \\
16.retrieve all speaks which is mainly about how artificial intelligence could affect our lives and its lecturer \\
17.gather talks that mainly discuss climate change and its global impact, and provide all countries mentioned \\
18.retrieve all speaks which is mainly about how mental health influences our daily lives and extract all the institution name mentioned \\
19.find talks that analyze the future of work in an automated world, and return the occupation of the lecturer \\
20.get all talks that address breaking gender stereotypes in modern society, and include the lecturer \\
21.retrieve all texts which is mainly about medicine, and extract what the lecturer will talk about \\
22.retrieve all texts which are mainly about health, and extract all the disease and its associated cause \\
23.find all texts which are mainly about literature, and extract all the awards of [PERSON] \\
24.find all texts which are mainly about science, and extract the profession of [PERSON] \\
25.output all texts which are mainly about history, and extract all the events and the time of the events \\
26.retrieve all speaks which is mainly about both health and brain in the speak, then extract their lecturer \\
27.retrieve all speaks which is mainly about both design and creativity in the speak, then extract all artists mentioned \\
28.retrieve all speaks which is mainly about both medicine and surgery in the speak, then extract all countries mentioned \\
29.retrieve all speaks which is mainly about artificial intelligence and ethics in the speak, then extract all location mentioned \\
30.retrieve all speaks which is mainly about artificial intelligence and machine learning in the speak, then extract the lecturer \\
31.gather all texts which is mainly about finance or artificial intelligence, and extract the lecturer \\
32.get all texts which is mainly about education or biology, and extract all professions  \\
33.return all texts which is mainly about philosophy or literature, and extract all person mentioned \\
34.output all speaks which centers on literature or philosophy, then extract all the university affiliation \\
35.retrieve all speaks which centers on music or visual arts, then extract the awards \\
36.retrieve all speaks which is mainly about health but is not about brain, then extract their lecturer \\
37.retrieve all texts which is mainly about environment but is not about climate change, and extract the locations \\
38.identify all talks mainly focusing on finance but not mentioning technology, then extract all lecturer name metioned \\
39.find all speeches mainly about artificial intelligence but without any reference to machine learning, then list all researchers mentioned \\
40.filter all talks centered on technological innovation but not mentioning blockchain, and extract all numbers mentioned \\

\end{botbox}



\end{document}